\documentclass[conference]{IEEEtran}
\IEEEoverridecommandlockouts
\usepackage{cite}
\usepackage{amsmath,amssymb,amsfonts,comment}
\usepackage{algorithmic}
\usepackage{graphicx}
\usepackage{textcomp}

\providecommand{\tightlist}{%
  \setlength{\itemsep}{0pt}\setlength{\parskip}{0pt}}

\newcommand\Tau{\mathrm{T}}

\def\BibTeX{{\rm B\kern-.05em{\sc i\kern-.025em b}\kern-.08em
    T\kern-.1667em\lower.7ex\hbox{E}\kern-.125emX}}
\begin{document}

\renewcommand{\baselinestretch}{0.98}

\title{A Grammar for the Representation of Unmanned Aerial Vehicles with 3D Topologies\\
}



\author{
\IEEEauthorblockN{Piergiuseppe Mallozzi$^1$, Hussein Sibai$^2$, Inigo Incer$^{1}$, Sanjit A. Seshia$^1$, Alberto Sangiovanni-Vincentelli$^1$}
\IEEEauthorblockA{$^1$ Electrical Engineering and Computer Sciences, UC Berkeley, USA}
\IEEEauthorblockA{$^2$ Computer Science and Engineering, Washington University in St. Louis, USA}
\IEEEauthorblockA{mallozzi,inigo,sseshia,alberto@berkeley.edu, sibai@wustl.edu}
}

\maketitle

\begin{abstract}
We propose a context-sensitive grammar for the systematic exploration of the design space of the topology of 3D robots, particularly unmanned aerial vehicles. It defines production rules for adding components to an incomplete design topology modeled over a 3D grid. The rules are local. The grammar is simple, yet capable of modeling most existing UAVs as well as novel ones. It can be easily generalized to other robotic platforms. It can be thought of as a building block for any design exploration and optimization algorithm.   
\end{abstract}

\section{Introduction}

Systematic exploration of a design space is fundamental to cyber-physical systems (CPS) design, yet many current methods are mainly manual and tedious.
It is common for designs to be optimized versions of previous ones as a result of {\em exploitation} of existing designs rather than {\em exploration} of the design space.   
An algorithm that automates the exploration process should be first able to generate {\em valid} designs. 
Formal languages have been essential for defining {\em valid} programs or specifications, usually as context-free or context-sensitive grammars. 
Recently, they have also been used to define {\em valid} 2D-shaped ground and aerial robots~\cite{xu2021moghs,robo-grammar-2020}. Designing 3D-shaped robots is significantly more challenging than 2D-shaped ones. Other proposed methods for design topology generation include using evolutionary algorithms~\cite{evolutionary_algorithms_for_design_mangharam_2022} and translating human-drawn sketches~\cite{trinity_SRI_2022}. 

In this abstract, we introduce a grammar to represent 3D-shaped designs of
Unmanned Aerial Vehicles (UAVs) under a set of design
rules. Our primary objective is to produce a 3D grid consisting of
interconnected points that represent the abstract topology of the UAV.
Each point within the grid represents a specific component of the UAV,
such as a fuselage, wing, or rotor, while each edge between two points
depicts a physical connection among components. To achieve this, we employ a
context-sensitive grammar (CSG) that formalizes the placement of UAV
components on the grid based on their local context, i.e., the
components around them.
Our grammar's primary goal is to generate all legal configurations of
components and their connections within the 3D grid. 

\section{A grammar for generating UAV topologies}
\label{topology-generation-for-uav-designs}

A context-sensitive
grammar (CSG) is a formal grammar that describes a formal language
through a set of rules that define how strings of symbols within the
language can be generated.
In this section, we introduce the main building blocks of our grammar:
{\em grid}, {\em context}, {\em symbols}, and {\em rules}. 
By leveraging these components, our
proposed method generates UAV designs that adhere to predefined design
rules, resulting in functional and efficient UAV designs.

\paragraph{Grid}\label{grid}
A 3D grid of points, or simply a grid, can be formally defined as a set 
of equally-spaced points along axis-parallel planes in the 3D space.
Each point is a triplet composed of three integer coordinates \((x, y, z)\) that
specify the position of the point in the 3D space. Each unit in the coordinate system corresponds to a physical Euclidean distance that is a parameter of the design algorithm that is using the grammar. Let
\(n_{\mathit{half}} \in \mathbb{Z}\), and let
$G = \{{(x, y, z) \in \mathbb{Z}^3 \;|\; -n_{\mathit{half}} \leq x, y, z \leq n_{\mathit{half}}}\}$
be the set of all points in a 3D grid. The size of the  grid
 is \(G\) as \(|G| = ((n_{\mathit{half}} * 2) + 1)^3\).



Using the defined grid structure, we  specify the position of
each component of the UAV in the form of a 3D point. 

\paragraph{Symbols}\label{symbols}

In our grammar, each point \(p\) in the grid is associated with a
symbol, denoted by \(Sym(p)\). The symbols are divided into two main
categories: {\em terminal} symbols and {\em nonterminal} symbols. Terminal symbols
represent actual characters or words in the language that the grammar
defines, while nonterminal symbols are used to represent groups or
structures in the language. These symbols can be expanded or rewritten
by the production rules of the grammar to generate strings in the
language.

We define the set of terminal symbols as $\Tau$, which contains
five elements that mostly represent UAV components:
\texttt{Fuselage}, \texttt{Rotor}, \texttt{Wing}, \texttt{Connector}, and \texttt{Empty}.
We define the set of nonterminal symbols as \(\Theta\), which contains
a single  element: \({\texttt{Unoccupied}}\).
Although \(\texttt{Empty}\) and \(\texttt{Unoccupied}\)
represent the same concept (i.e., the absence of a component), they are
different symbols. \(\texttt{Empty}\) is terminal, meaning it cannot be
expanded into a new symbol, while \(\texttt{Unoccupied}\) is
nonterminal.



\paragraph{Context and State}\label{context}

Given a 3D grid of points, we define the \textit{state} of a point \(p\), denoted by $S(p)$, as a tuple of seven symbols. Six of the entries correspond to the symbols of the points adjacent to \(p\) and one is the symbol associated with \(p\). Similarly a \textit{context} is a tuple of seven symbols but it is not linked to a specific state. We define a set of seven directions
\(\Delta = (\texttt{ego}, \texttt{front}, \texttt{rear}, \texttt{left}, \texttt{right}, \texttt{top}, \texttt{bottom})\).





\begin{figure*}[h]
  \centering
\includegraphics[width=0.85\textwidth]{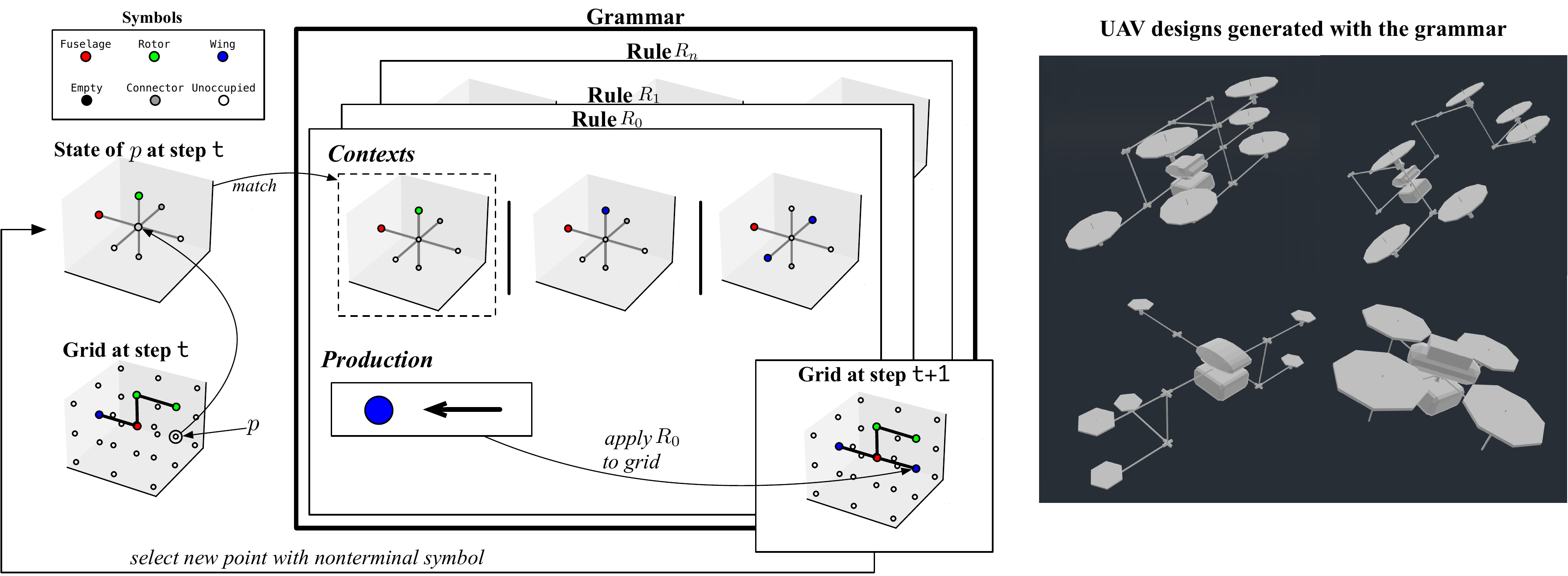}
  \caption{Design generation process using the context-sensitive grammar.}
  \label{fig:grammar}
  \vspace{-10px}
\end{figure*}









\paragraph{Rules}\label{rules}

In our grammar, a production rule, or rule, defines how a single
non-terminal symbol can be substituted with a terminal symbol
according to the state of the point associated with them. A rule \(R_i = (\Omega_i, \Pi_i)\) is a tuple of contexts and production where:

\begin{enumerate}
\def\labelenumi{\arabic{enumi}.}
\tightlist
\item
  \(\Omega_i=\{c_o, c_1, ..., c_n\}\) is a set of \(n\) contexts.
\item
  \(\Pi_i = (sym, \texttt{dir})\) where \(sym \in \Tau\) and
  \(\texttt{dir} \in \Delta\).
\end{enumerate}

\(\Omega_i\) defines all the contexts where the rule can be applied. The production \(\Pi\) specifies the nonterminal symbol \(sym \in \Tau\) to be produced and the direction of the edge that connects the newly generated symbol to one of the adjacent symbols in the grid. If no connection is required, then \(\texttt{dir}\) is simply
\(\texttt{ego}\).

\paragraph{Matching}\label{matching}

At each time step, we apply a rule to a point in the grid. A rule
\(R_i\) \emph{matches} with a point \(p\) if the state of \(p\),
i.e.,~\(S(p)\), belongs to the contexts of \(R_i\), i.e.,~if
\(S(p) \subseteq \Omega\). If a rule \(R_i = (\Omega_i, \Pi_i)\) matches
with a point \(p\), then the production \(\Pi_i\) can be applied to the state \(S(p)\). That is the symbol in \(\Pi_i\) can be placed at the point $p$ on the grid and a connection can be created between $p$ and the point corresponding to the adjacent symbol prescribed by $\Pi_i$.



\paragraph{Design generation
process}\label{grammar-and-design-generation-process}

Figure~\ref{fig:grammar} shows the matching process of state to rules of the grammar. We start with a grid where all points are associated with the
nonterminal symbol \texttt{Unoccupied}. Then, at discrete time-steps, we
choose a point in the grid and get all the rules in
the grammar that match it. 
We choose one of the matching rules according to a heuristic and apply the production of the rule to the point, which changes its
nonterminal symbol and potentially creates edges between adjacent
points. We repeat this process until all points contain terminal
symbols or until we have no rules matching in the grammar.

\section{Discussion}
The formalization of a grammar to encode design rules reduces the design space and improves the search for valid designs. However, this approach has trade-offs.

\emph{Rules origin}. We can automatically generate rules by analyzing existing designs. While we can augment the automatically generated rules with domain knowledge, there is no guarantee of completeness on all possible rules or correctness of the generated rules. Therefore, we need a feedback mechanism to identify and remove rules that generate incorrect designs.

\emph{Rule choice} and \emph{state exploration}. Given that states can be compatible with multiple rules, choosing the right rule to apply and the next state to visit can greatly affect the final topology of the design and the performance of the UAV. One can use reinforcement learning techniques to incorporate feedback from evaluation results, which can be used to update probability distributions on the choice of rule and next state to visit, improving the efficiency of the design search process.

\emph{Satisfying specifications}. A design is defined by the sequence of states visited and the rules applied at each state. We can view the design process as a transition system, where each transition indicates the rule applied and the next state to visit. To ensure that the resulting design satisfies certain specifications, we can formulate those specifications as constraints and employ synthesis or satisfiability techniques to generate a valid design that satisfies them. For example, we can encode the desired design specifications as an satisfiability modulo theories (SMT)~\cite{barrett-smtbookch21} formula and use an SMT solver to check if a given design satisfies the constraints. 
We have used Pacti~\cite{pacti} to formalize states and rules as assume-guarantee contracts using linear inequality constraints and employed algebraic operations among contracts to identify matching rules. This approach allows us to leverage the power of contract-based design and automated reasoning to efficiently search the space of possible designs for valid configurations.

\section{Conclusions}
We presented a context-sensitive grammar for the design of 3D topologies of UAVs. Each rule in the grammar define the conditions under which a component can be added to an incomplete 3D design topology at a certain location on a predefined 3D grid as well as how it would be physically connected to the rest of the design. The grammar is a building block for design space exploration using off-the-shelf tools such as reinforcement learning algorithms and SMT solvers. We were able to generate thousands of random valid 3D topologies of UAVs within few seconds. 
The grammar can be generalized for non-orthogonal topologies and account for more general constraints such as (hyper-)edges corresponding to non-physical connections or multi-hop neighbors. 


\bibliographystyle{abbrv}
\bibliography{ref}


\end{document}